\documentclass[runningheads]{llncs}

\usepackage{subfigure}

\usepackage{cite}
\usepackage{amsmath,amssymb,amsfonts}

\usepackage{graphicx}
\usepackage{textcomp}
\usepackage{xcolor}
\usepackage{algpseudocode}
\usepackage{algorithm}
\usepackage{float}
\usepackage[symbol]{footmisc}

\usepackage{tabularx}
\usepackage{caption}

\usepackage[T1]{fontenc}
\usepackage{graphicx}
\begin{document}
\title{Hierarchical Graph Neural Network with Cross-Attention for Cross-Device User Matching}
%
%
\author{Ali Taghibakhshi\inst{1,2,\thanks{This work was done while Ali Taghibakhshi was an intern at NVIDIA.}} \and
Mingyuan Ma\inst{1} \and
Ashwath Aithal\inst{1} \and
Onur Yilmaz\inst{1} \and
Haggai Maron\inst{1} \and
Matthew West\inst{2} 
}
\authorrunning{A. Taghibakhshi et al.}
%
\institute{NVIDIA \and
Department of Mechanical Science and Engineering, University of Illinois at Urbana-Champaign, Urbana, Illinois, USA}
\maketitle              
\begin{abstract}
Cross-device user matching is a critical problem in numerous domains, including advertising, recommender systems, and cybersecurity. It involves identifying and linking different devices belonging to the same person, utilizing sequence logs. Previous data mining techniques have struggled to address the long-range dependencies and higher-order connections between the logs. Recently, researchers have modeled this problem as a graph problem and proposed a two-tier graph contextual embedding (TGCE) neural network architecture, which outperforms previous methods. In this paper, we propose a novel hierarchical graph neural network architecture (HGNN), which has a more computationally efficient second level design than TGCE. Furthermore, we introduce a cross-attention (Cross-Att) mechanism in our model, which improves performance by 5\% compared to the state-of-the-art TGCE method.

\end{abstract}
\section{Introduction}

Recent years have witnessed significant flourish in engineering~\cite{gholizadeh2023evaluation, riera2014advances, gholizadeh2023suitability} Ensuring system security and effective data management are critical challenges in the modern day~\cite{gharaibeh2017smart, gholizadeh2016iec}. In this regard, data integration plays a vital role in facilitating data management, as it enables the integration of data from diverse sources to generate a unified view of the underlying domain. One of the primary challenges in data integration is the problem of entity resolution, which involves identifying and linking multiple data records that correspond to the same real-world entity. The problem of entity resolution arises in a wide range of domains, including healthcare, finance, social media, and e-commerce. Entity resolution is a challenging problem due to various factors, including the presence of noisy and ambiguous data, the lack of unique identifiers for entities, and the complexity of the relationships between different entities.

\begin{figure}[H]
\centering
\includegraphics[width=0.75\textwidth]{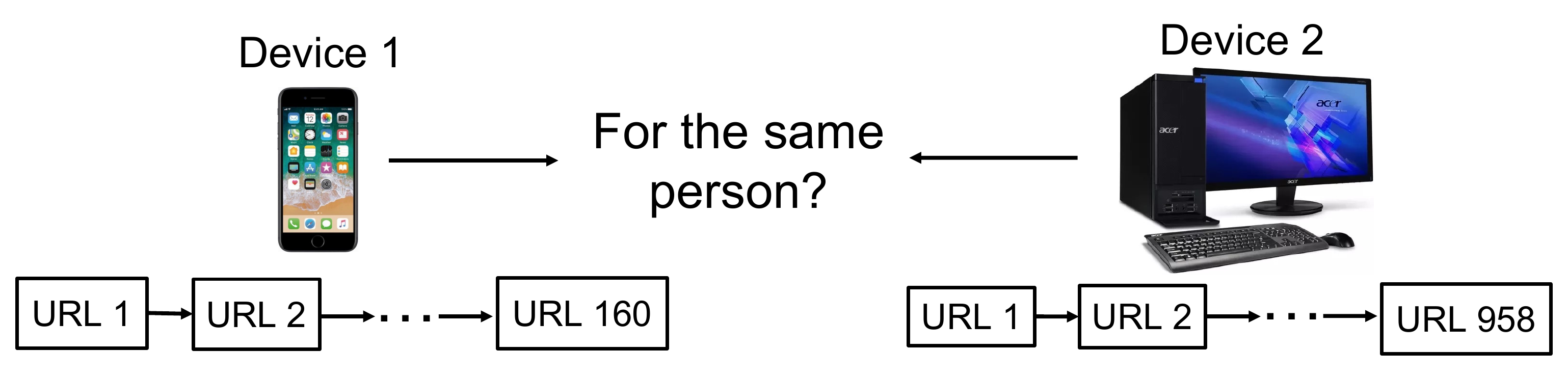}
\caption{Cross device user matching problem: only based the URL visit logs of two different devices, determine whether or not they belong to the same real-world person.}
\label{fig:cdum}
\end{figure}

Among entity resolution tasks, cross-device user matching is of significant importance. This task involves determining whether two separate devices belong to the same real-world person based on their sequential logs. The device sequential logs are time-stamped actions taken by the user over a relatively long period of time, say a few months. These actions are often in the form of browsing a Uniform Resource Locator (URL), and almost always, user identifications are not available due to privacy reasons. Refer to Figure~\ref{fig:cdum} for an illustration of the cross-device user matching task.

It is a common occurrence for users to engage in online activities across multiple devices. However, businesses and brands often struggle with having insufficient user identities to work with since users are perceived as different individuals across different devices due to their unique activities. The ability to automatically identify the same user across multiple devices is essential for gaining insights into human behavior patterns, which can aid in applications such as user profiling, online advertising, improving system security. Therefore, in recent years, the has been a flourishing amount of studies focusing on cross-device user matching~\cite{lian2016cross}.

In recent years, with the advent of machine learning-based methods for entity resolution, several studies have focused on learning distributed embeddings for the devices based on their URL logs~\cite{phan2016cross, phan2017cross, huang2021two}. The earlier studies focused on utilizing unsupervised feature learning techniques~\cite{le2014distributed}, developing handcrafted features for the device logs, or relied on co-occurrence of key attributes of URL logs in pairwise classification~\cite{phan2016cross}.

Methods that utilize deep learning have a greater ability to convey dense connections among the sequential device logs. For instance, researchers have utilized a 2D convolutional neural network (CNN) framework to encode sequential log representations to understand the relationship between two devices~\cite{tanielian2018siamese}. However, this model primarily captures local interactions within user sequence logs, limiting its ability to learn the entire sequence or a higher-level pattern. Recently, there has been further emphasis on the effectiveness of sequential models like recurrent neural networks (RNNs) and attention-based techniques in modeling sequence patterns and achieving promising results in numerous sequence modeling tasks~\cite{hidasi2015session, qiu2020gag, sun2019bert4rec}. Although these methods work well for sequence modeling, they are not specifically designed for user-matching tasks and may not be optimal for learning sequential log embeddings.

Recently, researchers proposed a two-tier graph contextual embedding (TGCE) network for the cross-device user matching~\cite{huang2021two} task. While previous methods for the task often failed at long-range information passing along the sequence logs, TGCE leverages a two-level structure that can facilitate information passing beyond the immediate neighborhood of a device log. This was specifically achieved by considering a random walk starting from every node in a device log, connecting all of the visited nodes to the original node, and performing a round of message passing using the newly generated shortcut edges.

Although the two-tier structure seems to enable long-range information sharing, we note two major limitations with the existing method. First, in the device graph, the random walk on the URL nodes may randomly connect two URLs that have been visited at two far-away time-stamps. Intuitively, two different URLs browsed by a device with weeks of gap in between share less information than two URLs visited in a shorter time frame. Second, at the end of the TGCE architecture, for the pairwise classification task, the generated graph embeddings for two devices are entry-wise multiplied and sent through a fully connected network to determine if they belong to the same person. However, there could be significant key features in the learned embeddings that may be shared between the devices, which can alternatively get lost if the architecture does not compare them across one another.

To address the above two issues, we propose a new hierarchical graph neural network (HGNN) inspired by the star graph architecture~\cite{pan2020star}. In the terminology of HGNN, we refer to the URL nodes as \textit{fine} nodes, and in an unraveled sequence of URL logs, HGNN assigns a \textit{coarse} node to every $K$ consecutive fine nodes. The message passing between the coarse and fine nodes enables effective long-range message passing without the need to excessively add edges, as in the random walk method. Moreover, for the pairwise classification task, we utilize a cross-attention mechanism inspired by Li \textit{et. al.}~\cite{li2020grapher}, which enables entry-wise cross-encoding of the learned embeddings. The main contributions of this paper are summarized as follows:
\begin{itemize}
\item We model a given device log as a hierarchical heterogeneous graph, which is 6x faster than the previous state-of-the-art while keeping a competitive level of accuracy and performance.
\item We employ a cross-attention mechanism for pairwise matching of the graphs associated with a device log, which improves the accuracy of the overall method by about 5\%.
\end{itemize}

\section{Related work}

The cross-device user matching task was first introduced in the CIKM Cup 2016\footnote{http://cikm2016.cs.iupui.edu/cikm-cup/}, and the first proposed methods for the task mainly considered hand-crafted features. For instance, the runner-up solution~\cite{lian2016cross} produces sub-categories based on the most significant URLs to generate detailed features. Furthermore, the competition winner solution proposed by Tay \textit{et. al.}~\cite{ramos2003using} utilizes ``term frequency inverse document frequency'' (TD-IDF) features of URLs and other related URL visit time features. However, their manually designed features did not fully investigate more intricate semantic details, such as the order of behavior sequences, which restricted their effectiveness. Aside from the hand-crafted features, the features that are developed from the structural information of the device URL visit data are also crucial for accomplishing the task of user matching. To further process sequential log information, studies have applied LSTM, 2D-CNN, and Doc2vec to generate semantic features for a sequence visited by a device ~\cite{tanielian2018siamese,phan2017cross,phan2016cross}.

Sequence-based machine learning models have also been employed for different entity resolution tasks; for instance, recurrent neural networks (RNN) have been utilized to encode behavior item sequential information~\cite{hidasi2015session}. Nevertheless, long-range dependencies and more advanced sequence features are not well obtained using sequence models~\cite{huang2021two}. With the advent of graph neural networks (GNN), researchers have modeled device logs as individual graphs where nodes and edges represent visited URLs and transitions between URLs. Each node and/or edge has an initial feature vector obtained from the underlying problem, and the layers of GNNs are then employed to update these features based on information passing in the local neighborhood of every node, such as the SR-GNN paper~\cite{wu2019session}. Another example is the LESSR~\cite{chen2020handling} method for recommendation systems where the method is capable of long-range information capturing using an edge-order preserving architecture. However, these methods are specifically designed for the recommendation task and do not necessarily achieve desirable results on the cross-device user matching task.

Recently, researchers have proposed TGCE~\cite{huang2021two}, a two-tier GNN for the cross-device user matching task. In the first tier, for every device log, each URL is considered as a node, and directional edges denote transitions between URLs. In the second tier, shortcut edges are formed by starting a random walk from every node and connecting all of the visited nodes to it. After a round of message passing in the first tier, the second tier is supposed to facilitate long-range information sharing in the device log. After the second tier, a position-aware graph attention layer is applied, followed by an attention pooling, which outputs the learned embedding for the whole graph. For the final pairwise classification, these learned embeddings for each of the devices are multiplied in an entry-wise manner and are sent to a fully connected deep neural network to determine whether they belong to the same user.

\section{Hierarchical graph neural network}

In this section, we discuss how we employ a two-level heterogeneous graph neural network for the cross-device user matching problem.

\subsection{Problem definition}
\label{subsec:probdef}

The aim of the cross-device user matching problem is to determine whether two devices belong to the same user, given only the URL visits of each device. Denote a sequence of visited URLs by a device $v$ by $\mathcal{S}_{v} = \{s_1, s_2, ..., s_n\}$, where $s_i$ denotes the $i$'th URL visit by the device (note that $s_i$'s are not necessarily different). We build a hierarchical heterogeneous graph, $G_v$, based on the sequence $\mathcal{S}_v$ as follows: for a visited URL, $s_i$, consider a \textit{fine} node in $G_v$ and denote it by $f_i$. Note that if multiple $s_i$'s correspond to the same URL, we only consider one node for it in $G_v$. Then, we connect nodes corresponding to consecutively visited URLs by directed edges in the graph; we connect $f_i$ and $f_{i+1}$ by a directional edge (if $f_i$ and $f_{i+1}$ correspond to the same URL, the edge becomes a self-loop). Up to this point, we have defined the fine-level graph, and we are ready to construct the second level, which we call the \textit{coarse} level.

To construct the second level, we partition the sequence $\mathcal{S}_{v}$ into non-overlapping subgroups of $K$ URLs, where each subgroup consists of consecutively visited URLs (the last subgroup may have less than $K$ URLs). For every subgroup $j$, we consider a coarse node, $c_j$, and connect it to all of the fine nodes corresponding to the URLs in subgroup $j$ via undirected edges.

\subsection{Fine level}
\label{subsec:fine-level}

In the fine level of the graph $G_v$, for every node $f_i$, we order the nodes corresponding to the URLs that have an incoming edge to $f_i$ according to their position in $\mathcal{S}_{v}$. We denote this ordered sequence of nodes by $N_{i} = \{f_{j_1}, f_{j_2}, ..., f_{j_{\kappa}}\}$. Also, we denote the feature vector of the fine node $f_i$ by $x_{i}$. The $l$-th round of message passing in the fine-level graph updates the node features according to the following update methods:
\begin{align}
& M_{i}^{(l)} = \Phi^{(l)}([x_{j_1}, x_{j_2}, ..., x_{j_\kappa}, x_i]), \\
& x^{(l+1)}_{i} = \Psi^{(l)}(x^{(l)}_{i}, M_{i}^{(l)}),
\end{align}
where $\Phi^{(l)}$ is a sequence aggregation function (such as sum, max, GRU, LSTM, etc.), for which we use GRU~\cite{chung2014empirical}, and $\Psi^{(l)}$ is a function for updating the feature vector (e.g., a neural network), for which we use a simple mean.

\subsection{Coarse level}
\label{subsec:corase-level}

\begin{figure}[!h]
\centering
\includegraphics[width=0.75\textwidth]{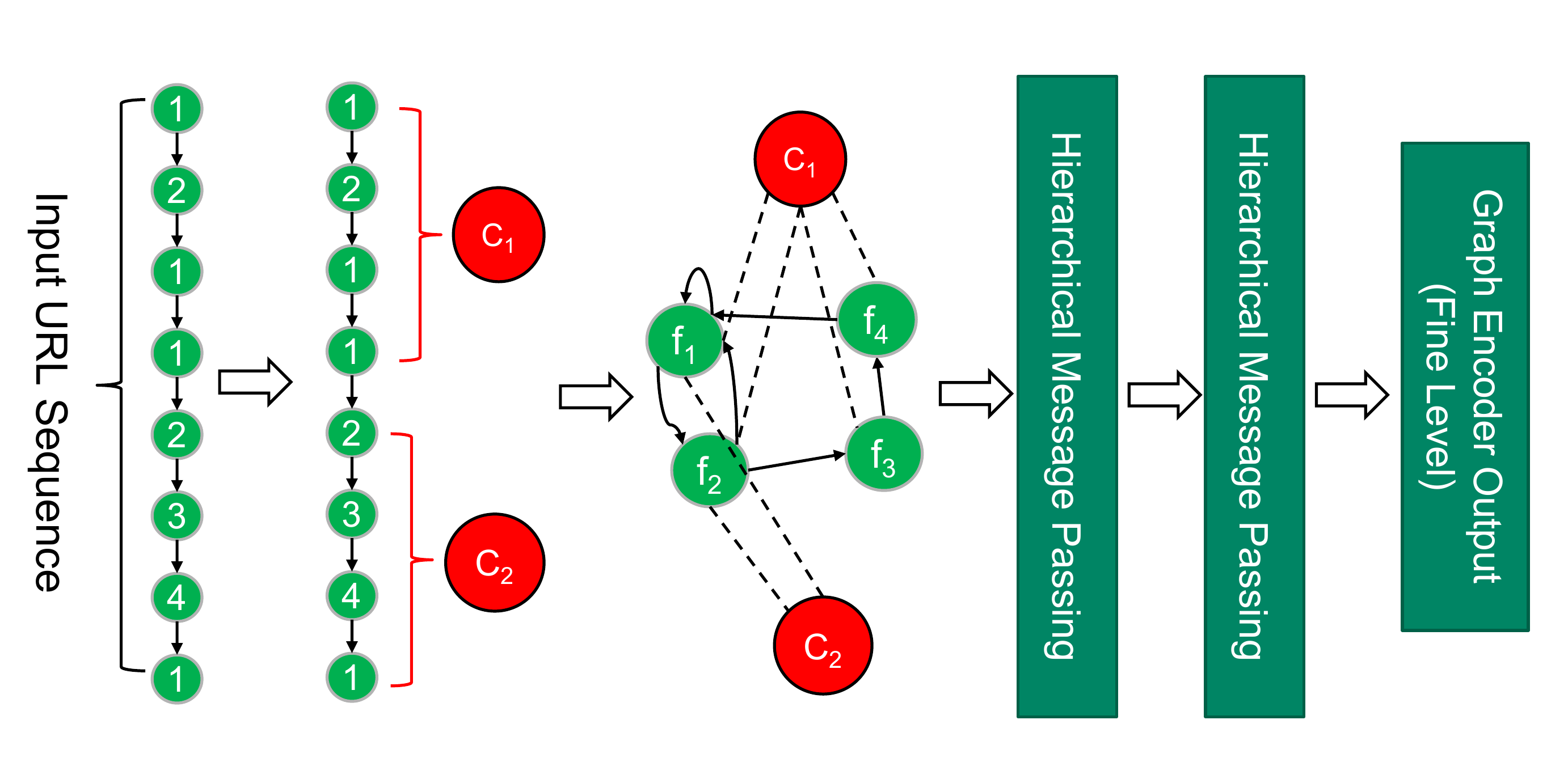}
\caption{From left to right: heterogeneous (fine and coarse) graph modeling from a given URL sequence. The hierarchical message passing blocks consist of message passing on the fine nodes with a GRU aggregation function. Next, the coarse node features are updated using a mean aggregation function. Finally, the fine node features are updated using their previous feature vector as well as an aggregated message from their associated coarse nodes obtained via an attention mechanism between coarse and fine level nodes.}
\label{fig:hgnn}
\end{figure}

In every round of heterogeneous message passing between fine and coarse level nodes, we update both the fine and coarse node features. Consider the coarse node $c_j$, and denote its feature by $\tilde{x}_j$. Also, denote the fine neighbor nodes of $c_j$ by $\mathcal{\tilde{N}}(c_j)$. In the $l$-th layer of heterogeneous message passing, the coarse node feature update is as follows:
\begin{equation}
\tilde{x}^{(l+1)}_j = \underset{{i}\in\mathcal{\tilde{N}}(c_j)}{\square}(W_{1}^{(i)}x_{i}),
\end{equation}
where $W^{(l)}_{1}$ is a learnable matrix and ${\square}$ is an aggregation function (such as mean, max, sum, etc.), for which we use mean. Denote by $\mathcal{{N}}(f_i)$ the set of coarse nodes connected to the fine node $f_i$. We first learn attention weights for the heterogeneous edges, and then we update fine nodes accordingly. In the $l$-th round of heterogeneous message passing, the fine node features are updated as follows:
\begin{align}
& e^{(l)}_{i,j} = \phi(W_{2}^{(l)}x^{(l)}_i, W_{3}^{(l)}\tilde{x}^{(l)}_j), \\
& \alpha^{(l)}_{i,j} = \frac{\text{exp}(e^{(l)}_{i,j})}{\sum_{j\in\mathcal{{N}}(f_i)}\text{exp}(e^{(l)}_{i,j})}, \\
& x^{(l+1)}_{i} = \xi(x^{(l)}_{i}, \sum_{j\in\mathcal{N}(f_i)}\alpha^{(l)}_{i,j}\tilde{x}^{(l)}_{j}),
\end{align}
where $W_{2}^{(l)}$ and $W_{3}^{(l)}$ are learnable matrices, and $\xi$ and $\phi$ are update functions (such as a fully connected network). Figure~\ref{fig:hgnn} shows the overall architecture of fine and coarse level message passing.

\subsection{Cross attention}
\label{subsec:cross-att}

After the message passing rounds in the fine level and long-range information sharing between fine and coarse nodes, we extract the learned fine node embeddings and proceed to cross encoding and feature filtering, inspired by the GraphER architecture~\cite{li2020grapher}. We consider two different device logs $v$ and $w$, and their learned fine node embeddings as a sequence, ignoring the underlying graph structure. We denote the learned fine node embeddings for device logs $v$ and $w$ as $X_v\in\mathbb{R}^{m_v\times d}$ and $X_w\in\mathbb{R}^{m_w\times d}$, where $m_v$ and $m_w$ are the number of nodes in the fine level of $G_v$ and $G_w$, respectively. We learn two matrices for cross-encoding $X_v$ into $X_w$ and vice-versa. Consider the $i$-th and $j$-th rows of $X_v$ and $X_w$, respectively, and denote them by $x_{v,i}$ and $x_{w,j}$. The entries $\hat{\alpha}_{i,j}$ of the matrix $A_{v,w}$ for cross-encoding $X_v$ into $X_w$ are obtained using an attention mechanism (and similarly for $A_{w,v}$):
\begin{align}
& \hat{e}_{i,j} = \zeta(W_{3}x_{v,i}, W_{3}x_{w,j}), \\
& \hat{\alpha}_{i,j} = \frac{\text{exp}(\hat{e}_{i,j})}{\sum_{k=1}^{m_w}\text{exp}(\hat{e}_{i,k})},
\end{align}
where $\zeta$ is an update function (such as a neural network), for which we use a simple mean. After obtaining the cross-encoding weights, we apply feature filtering, a self-attention mechanism that filters important features. The filtering vector is obtained as $\beta_{v} = \text{sigmoid}(W_{4} \text{tanh}(W_{5}X_v^{T}))$, where $W_{4}$ and $W_{5}$ are learnable weights ($\beta_{w}$ is obtained similarly). We apply the feature-filtering vector to the cross-encoding matrix as follows:
\begin{equation}
L_{v,w} = [\text{diag}(\beta_{v})(A_{v,w}X_{w} - X_{v})]\odot  [\text{diag}(\beta_{v})(A_{v,w}X_{w} - X_{v})],
\end{equation}
where $\odot$ denotes the Hadamard product ($L_{w,v}$ is also obtained similarly). The $L_{v,w}\in\mathbb{R}^{m_v\times d}$ and $L_{w,v}\in\mathbb{R}^{m_w\times d}$ matrices come from the Euclidean distance between the cross-encoding of $X_v$ into $X_w$ and $X_w$, and therefore are a measure of the closeness of the original sequence logs of $v$ and $w$.

To obtain a size-independent comparison metric, we apply a multi-layer perceptron (MLP) along the feature dimension of $L$ matrices (the second dimension, $d$), followed by a max-pooling operation along the first dimension. Finally, we apply a dropout and a ReLU nonlinearity. This yields vectors $r_{v,w}$ and $r_{w,v}$ that have a fixed size for any pair of $v$ and $w$. For the final pairwise classification task, we concatenate $r_{v,w}$ and $r_{w,v}$ and pass it through an MLP followed by a sigmoid activation to determine if the two devices belong to the same user or not:
\begin{equation}
\hat{y} = \text{sigmoid}(\text{MLP}(r_{v,w} || r_{w,v})).
\end{equation}

\begin{figure*}[!h]
\centering
\includegraphics[width=0.95\textwidth]{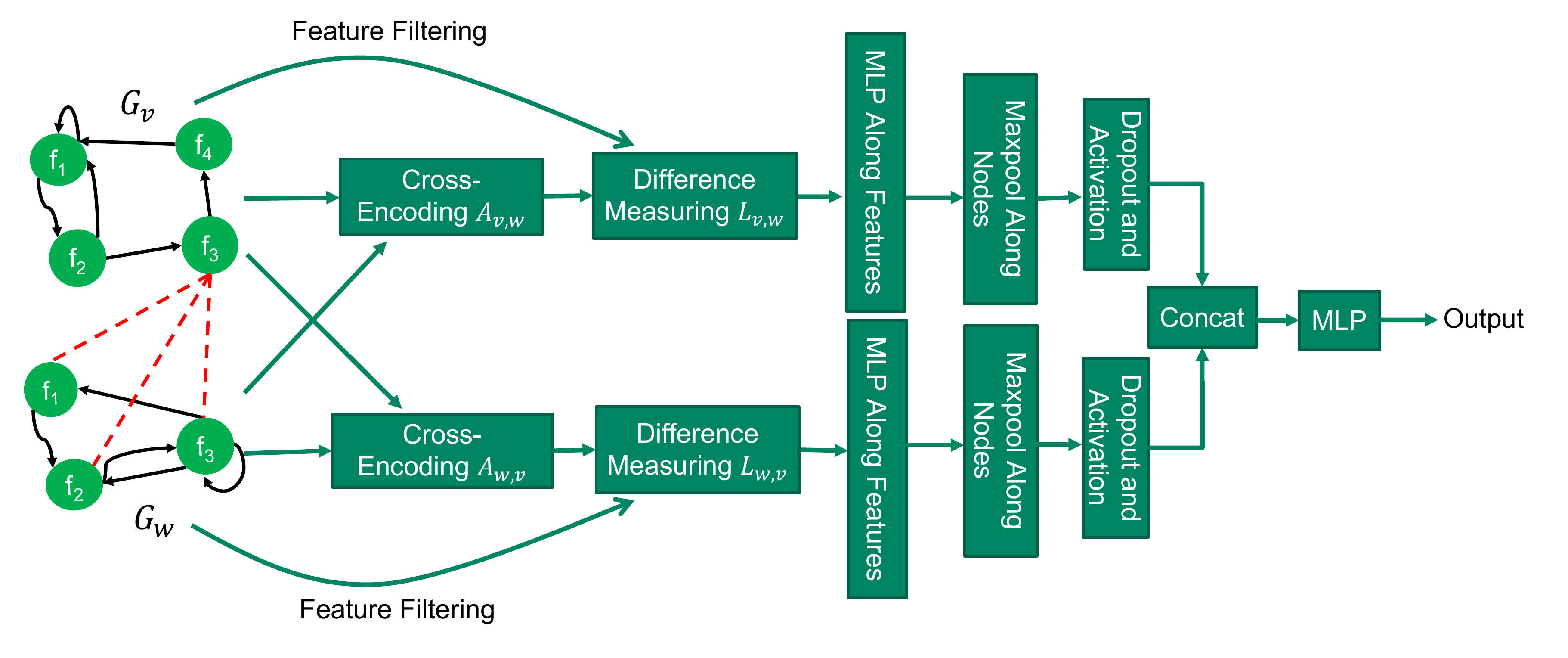}
\caption{Pairwise device graph matching: After the message passing, the two device graphs are cross-encoded via an attention mechanism followed by an attention-based feature filtering. The resulting matrix for each graph is then passed through an MLP layer, acting along the feature, followed by a maxpool operator along the nodes. Next, the obtained vectors pass through a dropout layer followed by an activation function. Finally, the resulting vectors of the two graphs are concatenated and passed through an MLP to obtain the final output.}
\label{fig:fig2}
\end{figure*}

\section{Experiment}
\label{sec:experiment}

In this section, we will describe the dataset, training details, and discuss how our method outperforms all other baselines, including TGCE~\cite{huang2021two}, the previous state-of-the-art.
\subsection{Training details}
\label{subsec:train}

\begin{figure}
\centering
\includegraphics[width=0.75\textwidth]{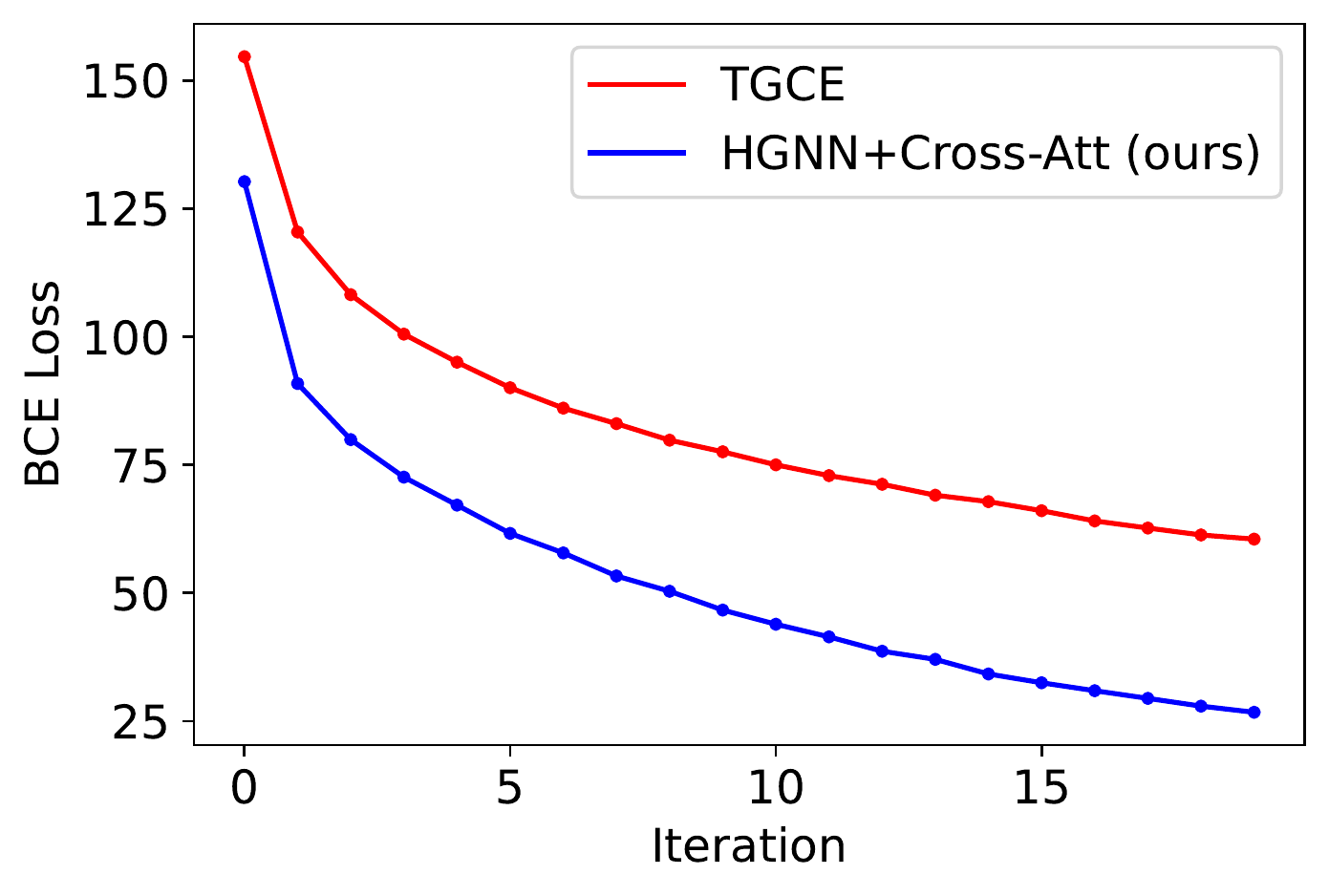}
\caption{Binary cross-entropy loss of our proposed method against that of TGCE. During training, our method obtains strictly better loss values.}
\label{fig:training}
\end{figure}

\begin{figure}
\centering
\includegraphics[width=0.75\textwidth]{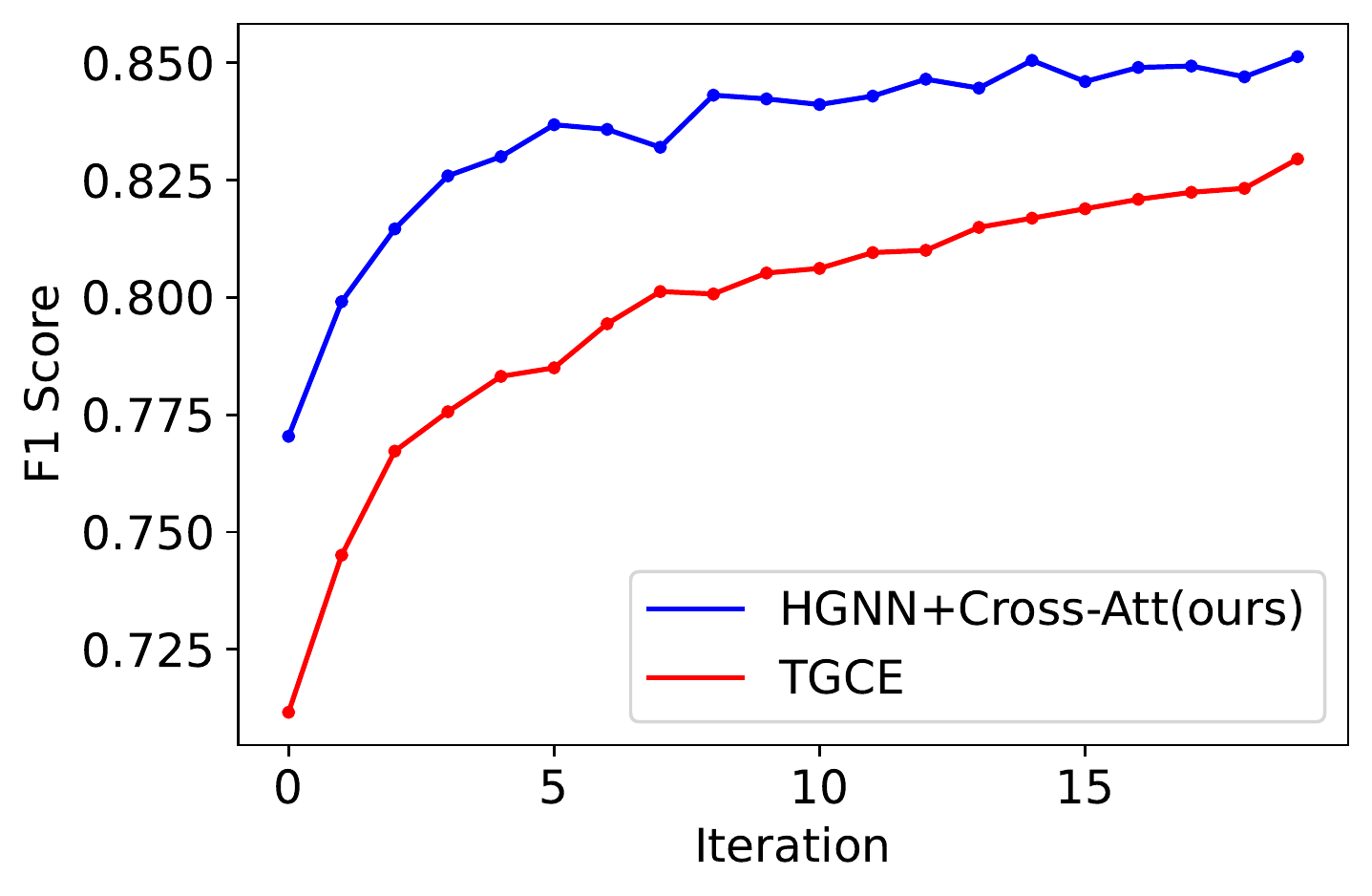}
\caption{Validation F1 score during training. Throughout the training, our method achieves strictly better F1 scores for the validation set compared to that of TGCE.}
\label{fig:validscore}
\end{figure}

We studied the cross-device user matching dataset made publicly available by the Data Centric Alliance\footnote{https://competitions.codalab.org/competitions/11171} for the CIKM Cup 2016 competition. The dataset consists of 14,148,535 anonymized URL logs of different devices with an average of 197 logs per device. The dataset is split into 50,146 and 48,122 training and test device logs, respectively. To obtain the initial embeddings of each URL, we applied the same data preprocessing methods as in~\cite{phan2017cross,huang2021two}. We used a coarse-to-fine node ratio of $k=6$, a batch size of 800 pairs of device logs, a learning rate of $10^{-3}$, and trained the model for 20 epochs. We used the binary cross-entropy (BCE) loss function for training our model. The training, evaluation, and test were all executed on an A100 NVIDIA GPU. The BCE loss during training as well as the validation F1 score are shown in Figures~\ref{fig:training} and~\ref{fig:validscore}, respectively.

\subsection{Results}
\label{subsec:results}

In this section, we evaluate the precision, recall, and F1 score of our method on the test set and compare it to available baselines. All of the baselines have been obtained similarly as described in~\cite{huang2021two}. We present two variants of our method; the first one, which we label ``HGNN'', only differs from TGCE in the design of the second tier, i.e., we use the hierarchical structure presented in subsections~\ref{subsec:fine-level} and~\ref{subsec:corase-level}, followed by the rest of the TGCE architecture. The second variant, which we label ``HGNN+Cross-Att'', uses the hierarchical structure in subsections~\ref{subsec:fine-level} and~\ref{subsec:corase-level}, and also utilizes the cross-attention mechanism presented in subsection~\ref{subsec:cross-att} after the hierarchical structure. As shown in Table~\ref{tab:f1scoreresult}, the ``HGNN+Cross-Att'' variant outperforms all of the baselines on the F1 score metric, including the second-best method (TGCE) by 5\% on the test data.

\begin{table}
\caption{Precision, Recall, and F1 score of different methods for cross-device user matching on DCA dataset.}
\begin{center}
    \begin{tabular}{|c|c|c|c|}
    \hline
    \textbf{}& \textbf{\begin{tabular}{@{}c@{}}Precision at Best\\ F1 Score\end{tabular}} & \textbf{\begin{tabular}{@{}c@{}}Recall at Best\\ F1 Score\end{tabular}}  & \textbf{\begin{tabular}{@{}c@{}} Best\\ F1 Score\end{tabular}} \\
    \hline
    TF-IDF & 0.33 & 0.27 & 0.26\\
    \hline
    Doc2vec & 0.29 & 0.21 & 0.24\\
    \hline
    SCEmNet & 0.38 & 0.44 & 0.41\\
    \hline
    GRU & 0.37 & \textbf{0.49}  & 0.42\\
    \hline
    Transformer & 0.39 & 0.47 & 0.43\\
    \hline
    SR-GNN & 0.35 & 0.34 & 0.34\\
    \hline
    LESSER & 0.41 & \textbf{0.48}  & 0.44\\
    \hline
    TGCE & 0.49 & 0.44 & 0.46\\
    \hline
    HGNN (ours) & 0.48 & 0.43 & 0.45\\
    \hline
    \begin{tabular}{@{}c@{}}HGNN+Cross-Att \\ (ours)\end{tabular}& \textbf{0.57} & \textbf{0.48} & \textbf{0.51}\\
    \hline
    \end{tabular}
\end{center}
\label{tab:f1scoreresult}
\end{table}

We also compare the training time of the two variants of our method with that of TGCE. As shown in Table~\ref{teb:timing}, our hierarchical structure is significantly more efficient than that of TGCE while keeping a competitive F1 score. Table~\ref{teb:timing} essentially indicates that by simply replacing the second-tier design of TGCE with our hierarchical structure (presented in subsections~\ref{subsec:fine-level} and~\ref{subsec:corase-level}), the method becomes 6x faster while almost keeping the same performance. This is due to the large number of artificial edges generated in the random walk passes in the creation of the second tier of TGCE. Moreover, although including cross-attention slows down the model, we can still obtain the same training time as TGCE and achieve 5\% better overall F1 score.

\begin{table}
\caption{Best F1 score and end-to-end training time of HGNN (without Cross-Att), HGNN+Cross-Att, and TGCE. The HGNN model is 6x faster than TGCE with a slight trade-off (about 1\%) on the accuracy side. The HGNN+Cross-Att model has the same training time as TGCE while achieving 5\% better F1 score.}
\begin{center}
    \begin{tabular}{|c|c|c|c|}
    \hline
    \textbf{}& \textbf{\begin{tabular}{@{}c@{}}Best \\ F1 Score\end{tabular}} & \textbf{\begin{tabular}{@{}c@{}}End-to-end\\ Training Time\end{tabular}} & \textbf{\begin{tabular}{@{}c@{}}Number of \\ Epochs\end{tabular}} \\ 
    \hline
   TGCE & 0.46 & 60h & 20\\
    \hline
    HGNN (ours) & 0.45 & \textbf{10h} & 20\\
    \hline
    \begin{tabular}{@{}c@{}}HGNN+Cross-Att \\ (ours)\end{tabular} & \textbf{0.51} & 60h & \textbf{6}\\
    \hline
    \end{tabular}
\end{center}
\label{teb:timing}
\end{table}

\begin{figure}[H]
\centering
\includegraphics[width=0.75\textwidth]{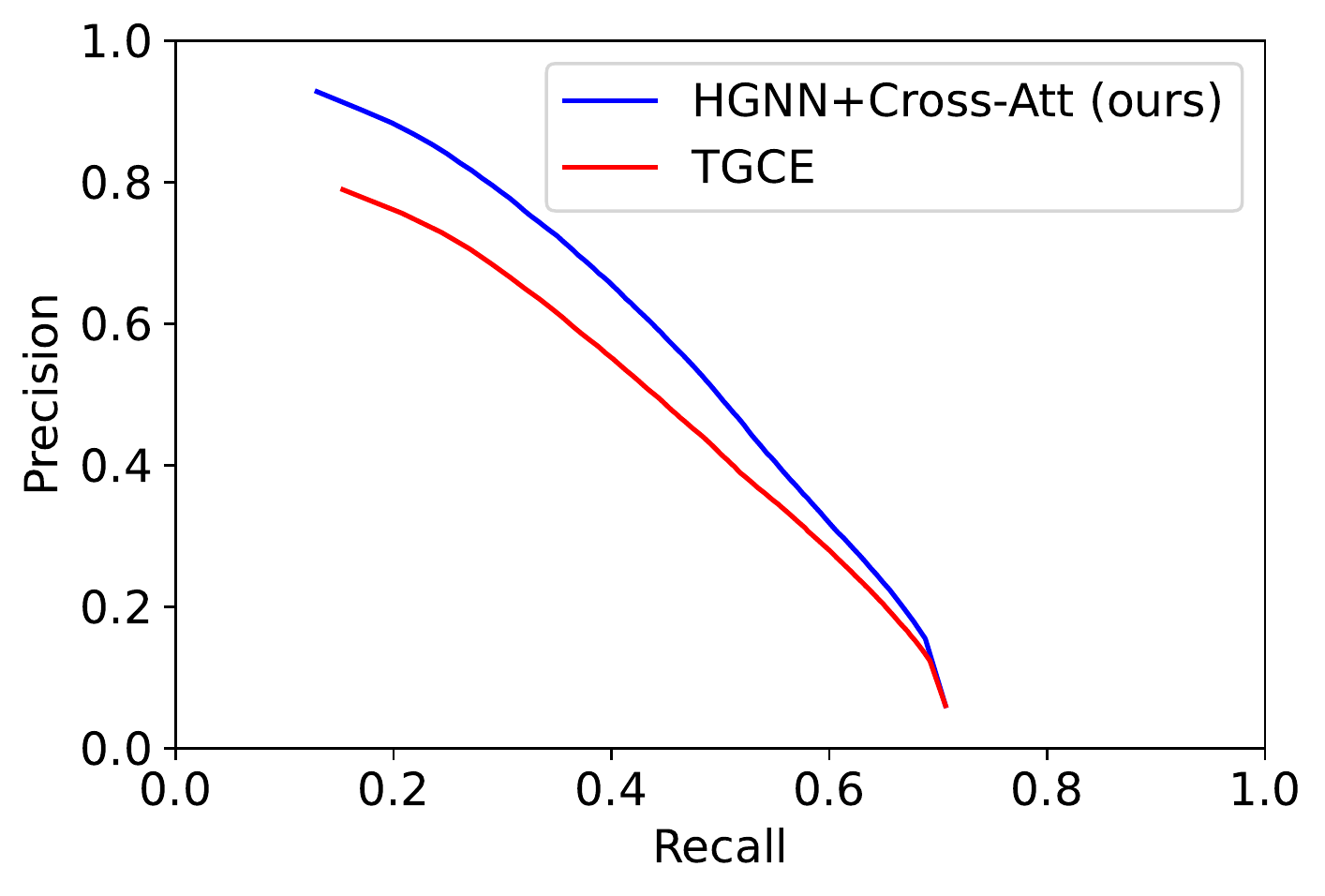}
\caption{Precision-Recall curve of the proposed method and that of TGCE on the test data.}
\label{fig:fplot}
\end{figure}

\begin{figure}[H]
\centering
\includegraphics[width=0.75\textwidth]{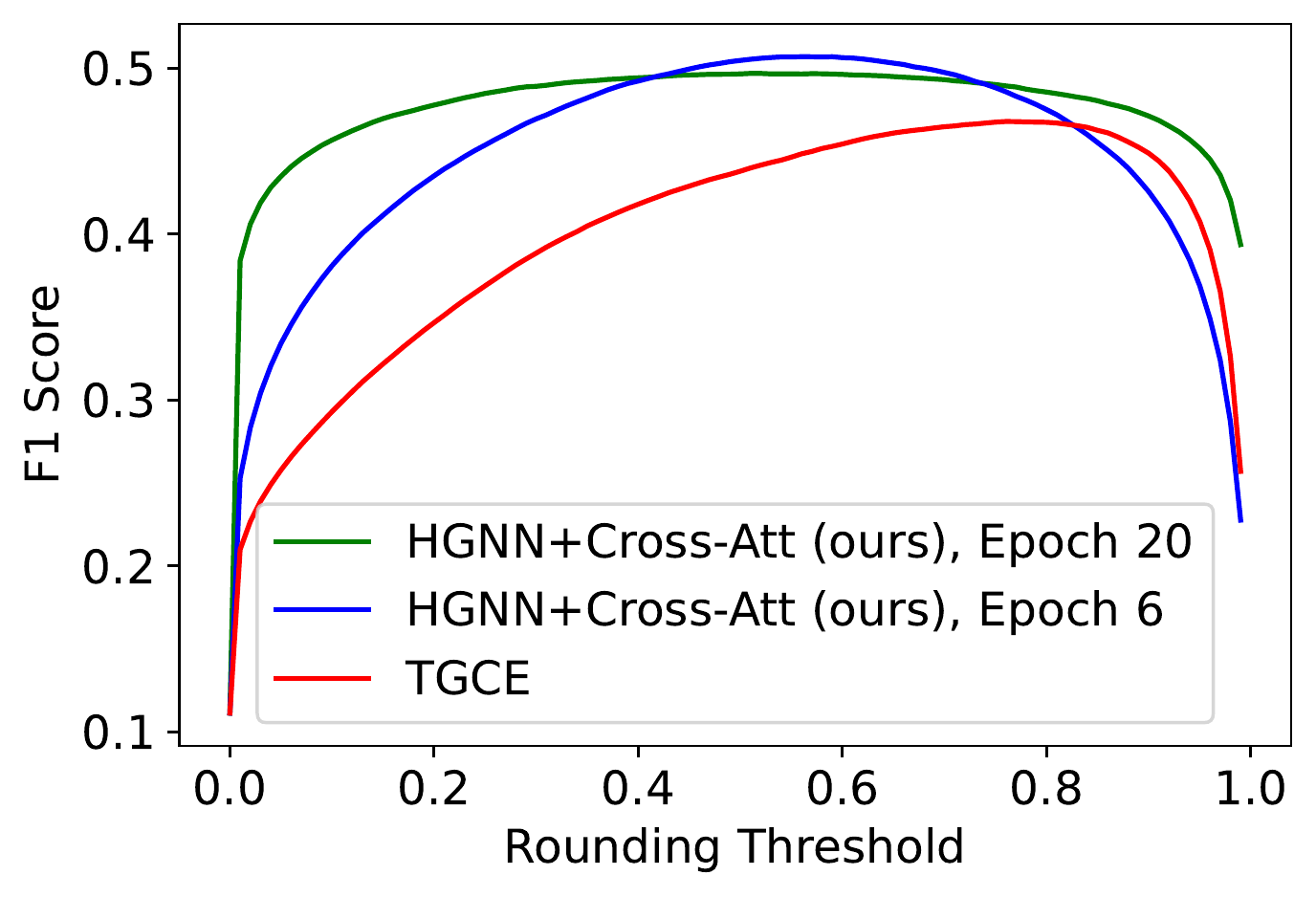}
\caption{F1 score against rounding threshold for our method (both for networks trained for 6 and 20 epochs) compared to that of TGCE (trained for 20 epochs). When our model is trained for 6 epochs, it achieves the highest overall F1 score, which is about 5\% better than that of TGCE. However, if our model is trained for 20 epochs, it achieves higher F1 scores for a wider range of rounding thresholds, outperforming TGCE for every rounding threshold.}
\label{fig:f1th}
\end{figure}

Figure~\ref{fig:fplot} shows the precision-recall curve of our method (the HGNN+Cross-Att variant, trained for 6 epochs) with that of TGCE (trained for 20 epochs). As shown in the figure, the precision-recall curve of our method is strictly better than that of TGCE. In other words, for every recall score, our method has a better precision. Additionally, we further trained the HGNN+Cross-Att variant for 20 epochs (the same number of epochs TGCE was trained for) to study if any further improvement is achieved on the test set. We also plot the F1 score with different thresholds (from 0 to 1 incremented by 0.01) for our model trained for 6 and 20 epochs and compare it to that of TGCE. As shown in Figure~\ref{fig:f1th}, our model trained for 20 epochs strictly outperforms TGCE (also trained for 20 epochs) for every threshold for obtaining the F1 score. However, our model trained for 6 epochs achieves the best overall F1 score, which is 5\% higher than TGCE. This is significant since as shown in Table~\ref{teb:timing}, the model takes the same time as TGCE to train.

\section{Conclusions}

In this paper, we present a novel graph neural network (GNN) architecture for a demanding entity resolution task: cross-device user matching, which determines if two devices belong to the same user based only on their anonymized internet logs. Our method comprises of designing an effective hierarchical structure for achieving long-range message passing in the graph obtained from device URL logs. After passing device logs through such a hierarchical GNN, we employ a cross-attention mechanism to effectively compare device logs against each other to determine if they belong to the same user. We demonstrate that our method outperforms available baselines by at least 5\%, while having the same training time as the previous state-of-the-art method, establishing the effectiveness of our proposed method.

\subsubsection{Acknowledgements} This research was supported by NVIDIA Corporation.

%
%
%
\bibliographystyle{splncs04}
\bibliography{hgnn_crossatt}

\end{document}